\documentclass[letterpaper]{article} 
\usepackage{aaai2027}  
\usepackage[hyphens]{url}  
\usepackage{graphicx} 
\usepackage{natbib}  
\usepackage{caption} 
\usepackage{algorithm}
\usepackage{algorithmic}
\usepackage{multirow}
\usepackage{graphicx} 
\usepackage{booktabs}
\usepackage{pifont}
\usepackage{bbding}
\usepackage{newfloat}
\usepackage{listings}

\usepackage[switch]{lineno}

\DeclareCaptionStyle{ruled}{labelfont=normalfont,labelsep=colon,strut=off} 
\floatstyle{ruled}
\newfloat{listing}{tb}{lst}{}
\floatname{listing}{Listing}

\usepackage{booktabs}

\nocopyright 

\title{Fine-Grained Anomaly Perception in Wild UGC-Enhanced Images: A Comprehensive Dataset and Difference-Fusion Framework}
\author{
    Yan Zhong\textsuperscript{\rm 1,2}\thanks{Yan Zhong and Gefei Chen did this work during their internship in ByteDance. This work is supported by the PKU-ByteDance \textit{Bad Case Detection} collaboration project.}, Gefei Chen\textsuperscript{\rm 1,2}, Qiufang Ma\textsuperscript{\rm 2}, Zhen Wang\textsuperscript{\rm 2}, Zhiwei Fan\textsuperscript{\rm 2}, Lei Shi\textsuperscript{\rm 3}, Tingting Jiang\textsuperscript{\rm 2}\corresponding\\
}
\affiliations{
    \textsuperscript{\rm 1} Peking University, Beijing, China\\
    \textsuperscript{\rm 2} Douyin, ByteDance Inc.\\
    \textsuperscript{\rm 3} Communication University of China, Beijing, China\\

    \{zhongyan@stu.,chengefei@stu.,ttjiang@\}pku.edu.cn; \{maqiufang.cv,wangzhen3560,fanzhiwei.rice\}@bytedance.com
}

\begin{document}

\maketitle

\begin{abstract}
Image enhancement and restoration have become standard back-end operations on short-video and social media platforms to boost UGC visual experience. Yet these processes inevitably introduce visual anomalies—especially in faces, texts, and textures—that directly undermine perceptual fidelity and viewer trust. While existing IQA methods perform well on classic distortions, they target holistic quality assessment and fail to capture the specific, localized anomalies caused by enhancement algorithms in real-world UGC.
To bridge this gap, we formally define a new task—quality Anomaly Perception for UGC image Enhancement (UEAP), and contribute the first UEAP benchmark dataset, named UEAP-4k, curated from the real business scenarios. It provides fine-grained annotations for anomaly categories, localization and severity levels.
Furthermore, we propose a Difference-Fusion Anomaly Perception Method (DFAP-UGC) for wild UGC-enhanced images, which leverages explicit problem-reference difference fusion with dense spatial querying, regional verification, and quality-aware ranking, enabling robust anomaly identification in challenging scenarios. 
To handle the inherent coupling of subtasks in this new task, we propose a Locality-Aware Dynamic Task Prioritization (LADTP) training strategy that enables effective end-to-end learning and eliminates multi-stage overhead.
Extensive experiments show that our method outperforms baselines adapted from classical approaches for this task, validating the value of this dataset and the superior of DFAP-UGC for robust UGC-enhanced image anomaly perception. Code and data will be public.
\end{abstract}


\section{Introduction}
With the rapid growth of short-video and social media platforms, user-generated content (UGC) has become a primary carrier of visual information~\cite{shao2019visual,zhong2025ctd}. To enhance visual experiences, platforms widely adopt quality enhancement and compression algorithms to automatically process UGC, with generative enhancement techniques further improving image clarity and color fidelity~\cite{safonov2025ntire}. However, such processing inevitably introduces local visual anomalies—such as content distortion, unintended recovery of private information, color artifacts, and texture irregularities, which severely undermine user experience and pose risks of complaints and brand damage~\cite{kushwaha2022overview}. Therefore, there is an urgent need for an anomaly perception approach tailored to UGC enhancement scenarios, capable of accurately identifying and localizing quality anomalies, thereby providing critical support for content quality assurance and algorithmic optimization.
\begin{table}[t]
\small
\centering
\resizebox{0.5\textwidth}{!}{
\begin{tabular}{l@{\hspace{0.4em}}l c c c c}
\toprule
\multicolumn{2}{l}{\textbf{Visual Quality Detection Tasks}} & \textbf{NR-IQA} & \textbf{FR-IQA} & \textbf{VAD} & \textbf{UEAP} \\
\midrule
\multirow{2}{*}{\textbf{Input}}
& Ideal Ref. Image & \ding{55} & \ding{51} & \ding{55} & \ding{55} \\
& Defective Ref. Image & \ding{55} & \ding{55} & \ding{55} & \ding{51}\\
& Test Image     & \ding{51} & \ding{51} & \ding{51} & \ding{51} \\
\midrule
\multirow{3}{*}{\textbf{Output}}
& Localization   & \ding{55} & \ding{55} & \ding{51} & \ding{51} \\
& Classification & \ding{55} & \ding{55} & \ding{55} & \ding{51} \\
& Regression (Score)    & \ding{51} & \ding{51} & \ding{55} & \ding{51} \\
\midrule
\multirow{2}{*}{\textbf{Aim}}
& Perceptual Distortion   & \ding{51} & \ding{51} & \ding{55} & \ding{51} \\
& Semantic Anomaly & \ding{55} & \ding{55} & \ding{51} & \ding{51} \\
\bottomrule
\end{tabular}
}
\caption{Comparison of the defined new task UEAP with other related visual quality detection tasks. NR-IQA and FR-IQA denote no-reference and full-reference image quality assessment for real-world distortion scenarios, respectively, while VAD stands for visual anomaly detection. Localization, Classification, and Regression in the output correspond to anomaly/distortion localization, anomaly/distortion categorization, and scoring of anomaly/distortion severity. The symbol \ding{51} indicates \textit{Yes} and \ding{55} indicates \textit{No}. These labels are defined for the majority of scenarios of each task. The proposed UEAP task requires no high-quality references but imposes stricter output criteria.
}
\label{tab1}
\end{table}

For anomaly perception in UGC-enhanced images, currently there are two lines of related work. \textbf{First}, Image Quality Assessment (IQA) is an important tool for evaluating image quality, including the quality of enhanced images. According to the availability of reference images, IQA methods are categorized as Full-Reference IQA (FR-IQA), Reduced-Reference(RR-IQA), No-Reference(NR-IQA)~\cite{yang2023deep}. For evaluating the quality of enhanced images, since usually there are no high quality reference images, many previous work adopt NR-IQA approach~\cite{TCSVT-2025,zhang2024task,chahine2025generalized,wang2026quality}
\textbf{Second}, visual anomaly detection (VAD)~\cite{cao2024survey,batzner2024efficientad} has also been widely explored: some VAD methods localize deviations by reconstruction errors~\cite{bergmann2019mvtec,zavrtanik2021draem}, feature distribution and memory-bank methods compare test patches with normal feature statistics~\cite{defard2021padim,roth2022total}, and transformer or distillation-based methods improve anomaly representation and localization~\cite{deng2022anomaly,liu2024deep}. 

However, neither IQA nor VAD can adequately perceive local anomalies in real-world UGC image enhancement scenarios. Specifically, 
(1)IQA methods only output global quality scores without fine-grained localization, classification and quality rating of 
abnormal regions,such as the artifacts of text, texture and human face caused by enhancement algorithms.
The FR setting requires the high quality reference image which is often missing in UGC scenarios.The NR setting ignores 
the pre-enhancement images which can provide important information for anomaly perception in UGC image enhancement.
(2) VAD methods localize deviations in the single test image, but they do not compare a post-enhancement image with a non-ideal pre-enhancement reference, making it difficult to distinguish newly introduced artifacts from original UGC defects or benign pairwise changes.

To fill this gap, this paper proposes a new image anomaly perception task, termed the fine-grained Anomaly Perception for image Enhancement in UGC scenarios (UEAP). As shown in Table~\ref{tab1}, this task differs from traditional IQA methods~\cite{mao2025no} that emphasize global quality scoring, and from generic visual anomaly detection~\cite{liu2024deep} tasks that identify abnormalities based solely on single-image distributions. Instead, our task targets pairwise local anomaly instance detection for generative UGC image enhancement: using the pre- and post-enhancement image pair as dual inputs, it accurately detects local unexpected bad cases introduced during enhancement, outputting bounding boxes, anomaly categories, and severity scores. It specifically focuses on high-risk local bad cases, such as text tampering, texture distortion, and abnormal portrait effect. 
Therefore, solving this task requires overcoming \textbf{three key Challenges: (i)} the lack of a real-world paired UGC enhancement benchmark with fine-grained and consistent annotations for local failure cases in short video scenarios; 
\textbf{(ii)} since the reference image is the pre-enhancement image rather than an ideal image, local anomalies are subtle and easily confounded with semantic variations, lacking clear discriminative cues.
\textbf{(iii)} jointly localizing, classifying, and grading anomalies is difficult due to their heterogeneous output spaces and the risk of inter-task interference.

To address \textbf{Challenge (i)}, we constructs the first dataset for this new task, drawn from public sources on video contents, termed the UGC-Enhanced Anomaly Perception (UEAP-4k) Dataset. Each sample consists of a pre-enhancement reference image, the corresponding post-enhancement image to be examined, and human-annotated labels for the latter, covering three aspects: anomaly localization, anomaly category, and severity score. To ensure more consistent and reliable annotations, we employ multiple annotators and propose a greedy merging strategy to reconcile their labels. The final dataset comprises over 4,000 samples, serving as the first benchmark for this task.

Based on UEAP-4k, we propose a Difference-Fusion Anomaly Perception method (named DFAP-UGC) for wild UGC-enhanced images, aiming to achieve robust anomaly identification in complex real-world scenarios.

To address \textbf{Challenge (ii)}, DFAP-UGC is designed around explicit pairwise comparison. As shown in Fig.~\ref{fig:method_overview}, it jointly uses the problem feature, the reference feature, and their difference, allowing the detector to focus on enhancement-induced changes rather than defects already present in the original UGC image. It further verifies candidate regions with local paired evidence and calibrates their ranking quality, providing compact yet discriminative cues for subtle text, texture, and portrait artifacts in complex UGC scenarios.

To address \textbf{Challenge (iii)}, considering the anomaly localization sub-task is the most challenging, we propose a Locality-Aware Dynamic Task Prioritization (LADTP) approach during training DFAP-UGC. Its core idea is to leverage Generalized Intersection over Union (GIoU) Loss\cite{rezatofighi2019generalized} as a real-time measure of localization quality, and jointly model it with the dynamical difficulty levels of sub-tasks to explicitly characterize task dependencies and adaptively adjust weights. Specifically, when localization performance is poor (high GIoU loss), the weights of classification and anomaly scoring losses are suppressed to focus the model on bounding box regression. As localization performance improves (low GIoU loss), other sub-tasks receive varying degrees of attention according to their respective difficulty levels, ultimately achieving balanced multi-task optimization. Additionally, LADTP updates weights with exponential moving average smoothing to ensure stable and robust training.


The main contributions are summarized below.
\begin{itemize}
    \item We formulate UEAP, a new fine-grained anomaly perception task for UGC image enhancement, and build UEAP-4k, the first benchmark dataset constructed from real-world short-video platforms. UEAP-4k contains paired pre-/post-enhancement images and unified annotations of ocalizing, classifying, and grading anomalies. Extensive experiments validating the value of this task and dataset.
    \item Tailored for this new task, we propose the first framework DFAP-UGC, which explicitly fuses problem-reference difference cues and combines dense spatial queries, regional verification, and quality-aware ranking to reliably identify enhancement-induced local anomalies.
    \item A Locality-Aware Dynamic Task Prioritization (LADTP) strategy is proposed to adaptively adjust sub-task weights based on localization quality and task difficulty, enabling balanced multi-task optimization during training.
\end{itemize}
\section{Related Works}
\subsection{Image Quality Assessment}
Image quality assessment (IQA) predicts perceptual image quality and is studied in full-, reduced-, and no-reference settings~\cite{zhai2020perceptual,mao2025no}. Full-reference methods compare a distorted image with a high-quality reference~\cite{lan2025hierarchical,liao2023full}, while no-reference methods are more practical for in-the-wild images where pristine references are unavailable~\cite{zhong2025semi,su2020blindly,feng2021error,zhong2025adaptive,sun2023graphiqa,zhao2023quality,zhong2024causal}.
Recent studies further exploit contrastive learning, semi-supervised representation learning, and vision-language or multimodal priors to improve generalization~\cite{saha2023re,prabhakaran2023image,wang2023exploring,zhang2023blind,DBLP:conf/icml/RadfordKHRGASAM21,chen2024promptiqa}.
Several efforts have specifically targeted enhanced images: a blind IQA model was learned from big data~\citep{TNNLS-2018}, a debiased subjective assessment protocol was proposed for real-world enhancement~\citep{CVPR-2021}, and more recently, a multi-scale feature fusion BIQA method with a large-scale natural image enhancement database (NIED) was introduced~\citep{TCSVT-2025}.
Meanwhile, AU-IQA~\cite{wang2025auiqa} provides a dedicated benchmark for AI-enhanced UGC, yet its annotations remain holistic quality scores, lacking instance-level anomaly localization and severity labels. This gap further motivates our UEAP task.
Overall, existing IQA methods still produce holistic quality scores and typically assume either no reference or an ideal reference. In contrast, UEAP takes a non-ideal pre-enhancement UGC image as reference and requires localizing enhancement-induced anomalies, classifying their types, and estimating severity---demanding explicit pairwise anomaly modeling.
\begin{figure}[t]
\centering
\includegraphics[width=0.48\textwidth]{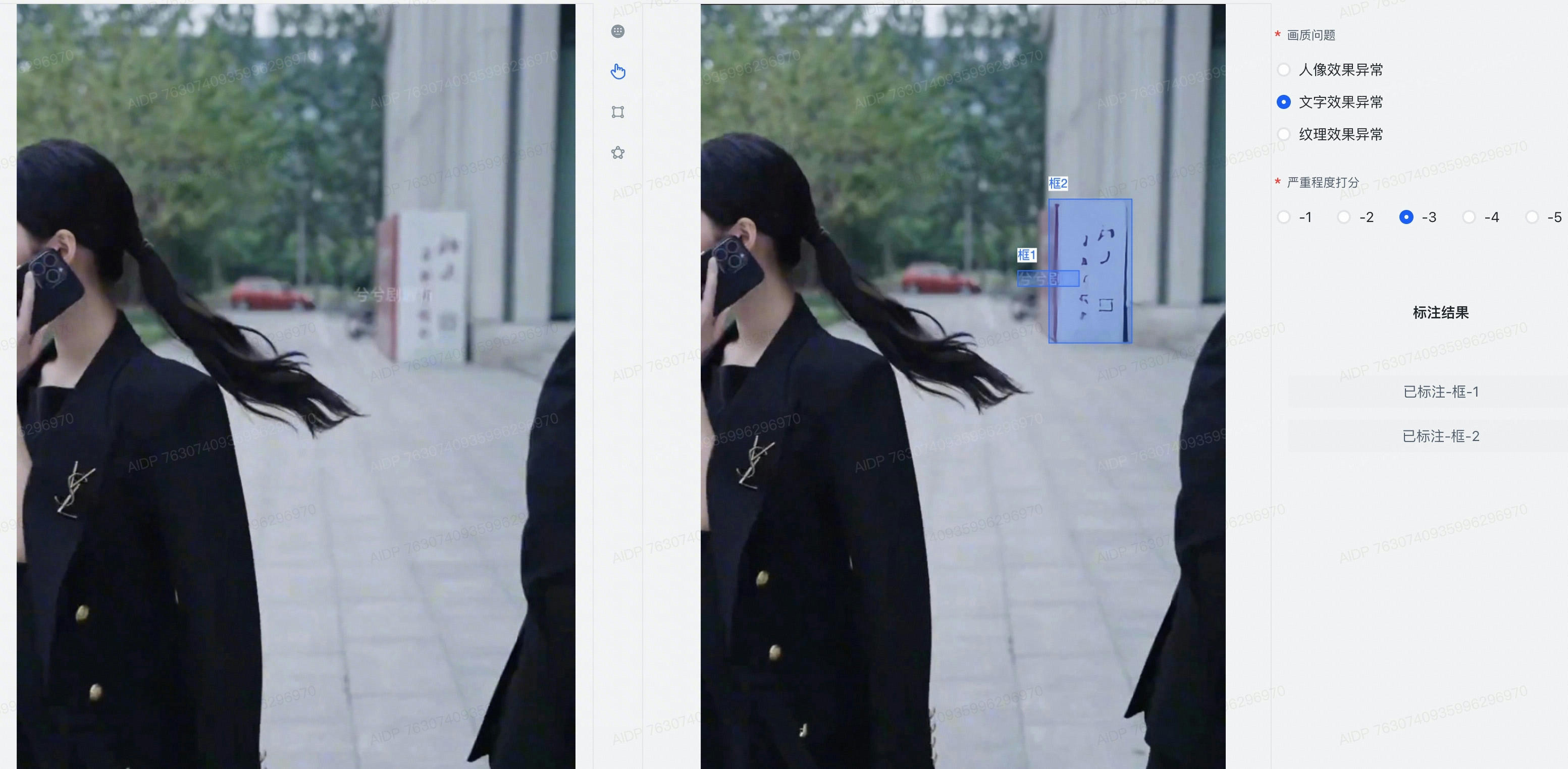}
\caption{Annotation interface and example for UEAP-4k. Annotators compare the reference and enhanced image side by side, draw bounding boxes on the enhanced image, select one anomaly type, and assign severity score from $-5$ to $-1$.}
\label{fig:annotation_panel}
\end{figure}
\subsection{Visual Anomaly Detection}
Visual anomaly detection (VAD) identifies deviations from normal visual distributions and has been widely studied for industrial inspection~\cite{liu2024deep,bergmann2019mvtec,zou2022spot}. Representative methods include reconstruction-based detection~\cite{zavrtanik2021draem,you2022unified}, patch-level distribution modeling and memory retrieval with pretrained features~\cite{defard2021padim,roth2022total}, and knowledge distillation based on teacher-student feature discrepancies~\cite{deng2022anomaly}. While effective in one-class or unsupervised settings~\cite{pang2021deep}, these methods judge a single image against normality to output anomaly scores or maps. UEAP instead compares a pre-enhancement UGC image with its enhanced counterpart and requires localization, category prediction, and severity estimation for enhancement-induced local failures. To address this setting, we contribute UEAP-4k, the first real-world benchmark, and propose DFAP-UGC, a tailored framework for robust pairwise anomaly perception in UGC enhancement scenarios.

\section{Task Definition and Datasets}
This section first formalizes the UEAP task and then describes the construction of the UEAP-4k benchmark.

\subsection{UEAP Task}
Given a pre-enhancement UGC image $I^r$ and its enhanced counterpart $I^e$, UGC image Enhancement Anomaly Perception (UEAP) aims to identify local quality failures newly introduced by the enhancement process. Different from full-reference IQA, whose reference is usually an ideal image and whose output is a global quality score, UEAP uses a non-ideal original UGC image as reference and focuses on instance-level abnormal changes. Formally, for each image pair $(I^r, I^e)$, the task outputs a set of anomaly instances $\mathcal{Y}=\{(b_i,c_i,s_i)\}_{i=1}^{N}$, where $b_i$ denotes the bounding box of an anomalous region in $I^e$, $c_i$ is its category, and $s_i \in [-5,-1]$ is the severity score. Following the annotation protocol, anomalies are grouped into three practical categories: portrait effect anomaly, text effect anomaly, and texture effect anomaly. The severity score measures how strongly the target image becomes worse than the reference, ranging from barely perceptible degradation to severe quality collapse or semantic corruption. The smaller the score is, the more severe the abnormality is. This formulation directly supports real UGC quality inspection by jointly requiring localization, categorization, and severity estimation.
\begin{figure}[t]
\centering
\includegraphics[width=0.48\textwidth]{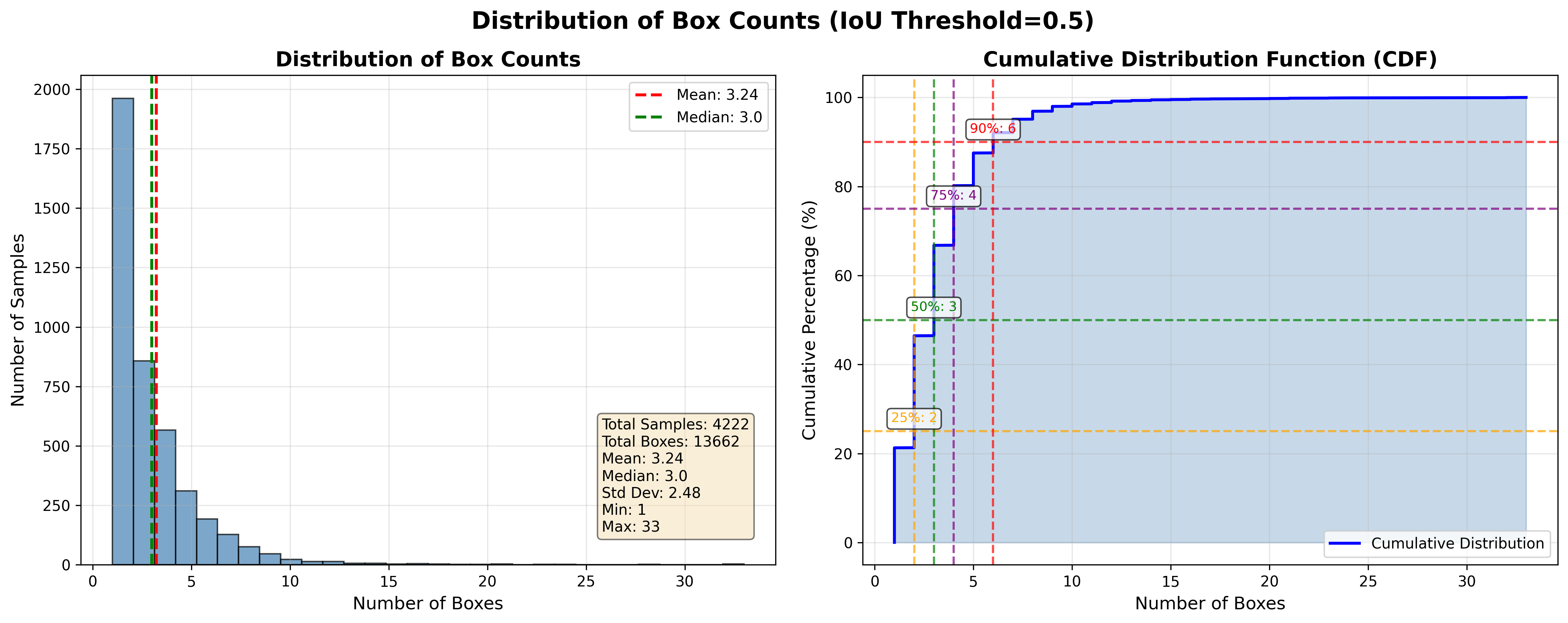}
\caption{Statistics of UEAP-4k after label unification. The left plot shows the histogram of anomaly box counts per sample, and the right plot shows the cumulative distribution.}
\label{fig:box_distribution}
\end{figure}
\begin{figure*}[t]
\centering
\includegraphics[width=\textwidth]{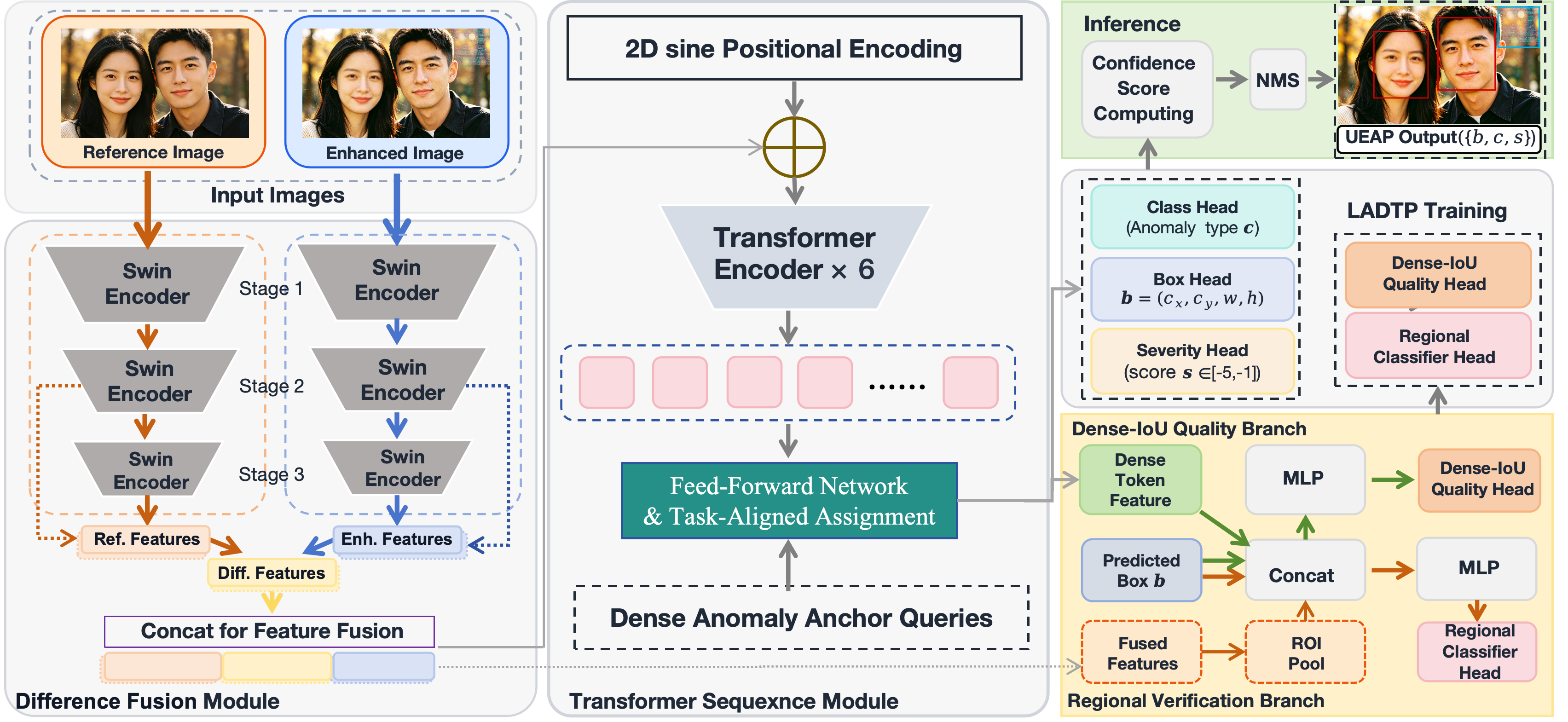}
\caption{Overview of our proposed Difference-Fusion Anomaly Perception Method (DFAP-UGC). 
}
\label{fig:method_overview}
\end{figure*}
\subsection{UEAP-4k Dataset}
\subsubsection{Data Source.} UEAP-4k is constructed from the public sources. 
We randomly sampled a subset of samples from these datasets and constructed the original reference images for the data used in this paper by randomly extracting frames. We then simulated a short-video platform pipeline, where the back-end system performs compression, encoding, decoding, and enhancement to reduce storage and bandwidth costs while improving playback quality, to obtain the enhanced images after processing. Although these operations are beneficial in most cases, they may also introduce subtle differences between the original uploaded content and the final displayed content, especially in human faces, text regions, and high-frequency textures. Such changes are often local and easily overlooked by global quality scores, but they directly affect user experience when they cause facial deformation, text corruption, texture hallucination, or semantic inconsistency. We therefore sample frames from the back-end video processing pipeline and collect paired images, where each pair consists of a pre-enhancement reference image and its corresponding post-enhancement target image. The goal is to annotate local bad cases that are introduced or amplified by the enhancement process.
\subsubsection{Annotation Protocol.} Following the task definition, each sample is annotated by three professional annotators using the same web-based annotation interface, as shown in Fig.~\ref{fig:annotation_panel}. The reference image and enhanced target image are displayed side by side, and annotators are instructed to mark only regions in the target image that become worse than the reference. To reduce missed cases and make the criterion consistent, annotation is conducted in three rounds, focusing on portrait effect anomalies, text effect anomalies, and texture effect anomalies. For each abnormal region, the annotator draws a bounding box, selects the anomaly category, and assigns a severity score in $[-5,-1]$. A score of $-1$ indicates a barely perceptible artifact or minor semantic change, while $-5$ indicates severe visual degradation or semantic corruption. Before formal labeling, all annotators are trained with the same guideline and example cases, and the annotation is performed on desktop displays under a consistent interface layout. During labeling, annotators are required to inspect the paired images independently, avoid marking regions whose differences are caused only by normal enhancement, and ensure that each box tightly covers the affected region.

\subsubsection{Label Unification.} Since different annotators may draw slightly different boxes or assign different scores to the same abnormal region, we design a greedy label unification strategy to merge multiple annotations into a single target label set. 
Firstly, the raw records are first grouped by sample identity. Then, for each sample, annotations are grouped by anomaly category, and boxes within the same category are greedily clustered according to an IoU threshold $\tau=0.5$. Matched boxes are merged by averaging their coordinates, severity scores are averaged, and the final textual label is determined by majority voting. Unmatched boxes are retained as individual instances, which helps preserve rare but valid local failures. The proposed greedy label unification strategy is summarized in Algorithm~\ref{alg:label_merge}.
\begin{algorithm}[t]
\caption{Greedy label unification for UEAP-4k}
\label{alg:label_merge}
\begin{algorithmic}[1]
\REQUIRE Raw annotation records $\mathcal{A}$, IoU threshold $\tau$
\ENSURE Unified dataset $\mathcal{D}$
\FOR{each sample group $G$}
    \STATE Collect valid boxes, categories, and severity scores from all annotators.
    \FOR{each anomaly category $c$}
        \STATE Put all boxes of category $c$ into queue $Q_c$.
        \WHILE{$Q_c$ is not empty}
            \STATE Pop an anchor annotation $a$ from $Q_c$.
            \STATE Find matches $\mathcal{M}=\{b\in Q_c\mid \mathrm{IoU}(a,b)>\tau\}$.
            \STATE Remove $\mathcal{M}$ from $Q_c$.
            \STATE Average box coordinates and severity scores of $\{a\}\cup\mathcal{M}$.
            \STATE Assign category $c$ and the majority textual label.
        \ENDWHILE
    \ENDFOR
    \STATE Add the merged annotations of $G$ to $\mathcal{D}$.
\ENDFOR
\RETURN $\mathcal{D}$
\end{algorithmic}
\end{algorithm}
\subsubsection{Dataset Statistics.} After label unification, UEAP-4k contains 4,222 paired samples and 13,662 anomaly boxes from 10,470 valid raw annotation records. The dataset covers three anomaly categories: portrait effect anomaly, texture effect anomaly, and text effect anomaly, with 7,921, 3,153, and 2,588 boxes, respectively. Portrait-related artifacts account for 58.0\% of all instances, indicating that face and body regions are particularly sensitive to enhancement failures. Texture and text anomalies account for 23.1\% and 18.9\%, respectively, reflecting complementary risks in high-frequency details and semantic symbols. Fig.~\ref{fig:box_distribution} further shows the distribution of box counts per sample. Each sample contains 3.24 boxes on average, with a median of 3, a standard deviation of 2.48, and a range from 1 to 33. The cumulative distribution shows that 75\% of samples contain no more than 4 boxes and 90\% contain no more than 6 boxes, while a small number of hard cases contain many local artifacts. This long-tailed distribution makes UEAP-4k realistic and challenging for robust local anomaly perception.
\section{Methodology}
\subsection{Overview}
Figure~\ref{fig:method_overview} shows our proposed \textbf{Difference-Fusion Anomaly Perception Method (DFAP-UGC)} for paired UGC enhancement anomaly perception. Given a pre-enhancement reference image $I^r$ and a post-enhancement problem image $I^p$, DFAP-UGC detects local enhancement-induced failures on the problem image and predicts their category, bounding box, and severity score. The method  explicitly presents the model with problem features, reference features, and their absolute difference, preserving a $28\times28$ detection grid. DFAP-UGC follows three coupled paths. The main detection path extracts aligned features with a shared Swin-T backbone~\cite{liu2021swin}, constructs explicit Stage-3 difference-fusion tokens, and uses a dense Transformer encoder~\cite{vaswani2017attention,carion2020end} to produce spatial detection queries. The regional path samples high-resolution Stage-2 paired features inside each predicted box to verify local evidence. The quality path estimates class-conditional localization quality from detached query and box features, which calibrates inference ranking. During training, we propose the Locality-Aware Dynamic Task Prioritization (LADTP), which can dynamically adjust the loss weights according to localization quality and task difficulty, achieving balanced multi-task optimization and robust anomaly perception.
\subsection{Shared Swin Encoder and Difference Fusion}
Both images are resized to $448\times448$ and processed by the same Swin-T encoder~\cite{liu2021swin}. Let $\mathbf{F}^p_s$ and $\mathbf{F}^r_s$ denote the projected enhanced and reference image features at Stage $s$. Under the input resolution, Stage 2 has a $56\times56$ grid and Stage 3 has a $28\times28$ grid. DFAP-UGC uses Stage 3 and Stage 2 for the main detection path and the regional verification path, respectively.

For the main path, the model builds a triplet difference representation at Stage 3:
\begin{equation}
\mathbf{D}_3=|\mathbf{F}^p_3-\mathbf{F}^r_3|,
\end{equation}
\begin{equation}
\mathbf{Z}=\mathrm{LN}\left(\mathrm{GELU}\left(W_f[\mathbf{F}^p_3;\mathbf{F}^r_3;\mathbf{D}_3]\right)\right),
\end{equation}
where $[\cdot;\cdot]$ denotes channel concatenation. This representation keeps the current visual content, the reference-side context, and the explicit local difference. The fused feature is then projected to the Transformer hidden dimension, producing $28\times28=784$ dense tokens.
\subsection{Dense Transformer Sequence Module}
The fused dense tokens are combined with 2D Sine positional encodings~\cite{carion2020end} and fed into a six-layer Transformer encoder:
\begin{equation}
\mathbf{H}=T_{enc}(\mathbf{Z}+\mathbf{P}),\quad \mathbf{H}=\{\mathbf{h}_i\}_{i=1}^{784}.
\end{equation}
Each output token $\mathbf{h}_i$ corresponds to a fixed grid position and is treated as a dense anchor query. For the query located at row $u$ and column $v$, its initial anchor center is $((v+0.5)/28,(u+0.5)/28)$, with default width and height set to 0.15. The box head predicts an anchor-relative delta:
\begin{equation}
\hat{\mathbf{b}}_i=\sigma(\mathrm{logit}(\mathbf{a}_i)+\mathrm{MLP}_b(\mathbf{h}_i)),
\end{equation}
where $\mathbf{a}_i$ is the fixed anchor. The category and severity predictions are
\begin{equation}
\hat{\mathbf{p}}_i=W_c\mathbf{h}_i,\quad
\hat{s}_i=-5\cdot\sigma(W_s\mathbf{h}_i).
\end{equation}
Here $\hat{\mathbf{p}}_i$ contains class logits for the three foreground anomaly types, $\hat{\mathbf{b}}_i$ is the normalized box in $(c_x,c_y,w,h)$ format, and $\hat{s}_i\in[-5,0]$ is the predicted severity score. The severity score is an output attribute and is not used as the detection confidence.
\subsection{Task-Aligned Assignment and Detection Losses}
Because DFAP-UGC uses 784 dense queries, one-to-one Hungarian matching is too sparse for training the neighborhood around each local anomaly. We therefore use dense task-aligned assignment (TAL)~\cite{Feng_2021_ICCV}. For query $i$ and ground-truth box $j$, the alignment score is
\begin{equation}
A_{ij}=p_i(y_j)^{\alpha}\cdot \mathrm{IoU}(\hat{\mathbf{b}}_i,\mathbf{b}_j)^{\beta},
\end{equation}
where $p_i(y_j)$ is the predicted probability of the ground-truth category. We use $\alpha=1$, $\beta=6$, and select the top-$k$ in-box queries for each ground-truth instance with $k=5$. If no query center lies inside a ground-truth box, the query closest to the box center is selected. This assignment produces multiple positives per anomaly and aligns classification confidence with localization quality.

The main classification loss is a dense focal-style loss~\cite{lin2017focal} over the TAL-selected positives and dense background queries:
\begin{equation}
\begin{array}{l}
\mathcal{L}_{cls}=\frac{1}{\sum_{i,c} y_{ic}}\sum_{i,c} w_{ic}\,\mathrm{BCE}(p_{ic},y_{ic}),\\
w_{ic}=y_{ic}+(1-\alpha_f)p_{ic}^{\gamma}\mathbf{1}[y_{ic}=0].
\end{array}
\end{equation}
Here $p_{ic}=\sigma(\hat{p}_{ic})$, $y_{ic}$ denotes the TAL target, i.e., the IoU score for selected positive query-class pairs and 0 otherwise, and $\gamma=2$. Localization and severity are optimized by
\begin{equation}
\mathcal{L}_{bbox}=\frac{1}{N}\sum\nolimits_i\|\hat{\mathbf{b}}_i-\mathbf{b}_i\|_1,
\mathcal{L}_{giou}=\frac{1}{N}\sum\nolimits_i(1-\mathrm{GIoU}(\hat{\mathbf{b}}_i,\mathbf{b}_i)),
\end{equation}
\begin{equation}
\mathcal{L}_{score}=\frac{1}{N}\sum\nolimits_i |\hat{s}_i-s_i|.
\end{equation}
During LADTP training, we use category-aware weights for localization-related positives to reduce the dominance of easier categories and emphasize harder texture and text failures.
\subsection{Regional Verification Branch}
Dense queries improve recall but can also create high-scoring false positives. The regional verification branch checks whether each predicted box contains local paired evidence for the predicted anomaly category. It uses the higher-resolution Stage-2 features. After projecting the problem and reference Stage-2 features to 64 channels, we construct
\begin{equation}
\mathbf{R}_2=[\mathbf{F}^p_2;\mathbf{F}^r_2;|\mathbf{F}^p_2-\mathbf{F}^r_2|].
\end{equation}
For each predicted box $\hat{\mathbf{b}}_i$, a $3\times3$ Region of Interest (RoI) feature is sampled from $\mathbf{R}_2$ by bilinear grid sampling. The regional classifier takes the dense query feature, the pooled regional feature, and the predicted box coordinates as input:
\begin{equation}
\hat{\mathbf{r}}_i=R([\mathbf{h}_i;\mathrm{Pool}(\mathbf{R}_2,\hat{\mathbf{b}}_i);\mathrm{sg}(\hat{\mathbf{b}}_i)]).
\end{equation}
Output $\hat{\mathbf{r}}_i$ is class-conditional regional logits. $\mathrm{sg}(\cdot)$ denotes stop-gradient. The branch is trained end-to-end with respect to the shared feature representation, while the sampling coordinates are detached so that the regional loss does not directly update the box head through the sampling operation. Hard negative mining is used in this branch to strengthen suppression of visually salient but non-anomalous regions.
The regional branch is optimized by a class-conditional binary cross-entropy loss:
\begin{equation}
\begin{array}{l}
\mathcal{L}_{reg}=\frac{1}{N_{reg}^{+}}\sum_{i,c}\omega^{reg}_{ic}\,\mathrm{BCE}(\sigma(\hat{r}_{ic}),y^{reg}_{ic}),\\
y^{reg}_{ic}=m^{reg}_{ic}\mathbf{1}[m^{reg}_{ic}\ge\tau_{reg}],
m^{reg}_{ic}=\max\limits_{\mathbf{b}\in\mathcal{B}_c}\mathrm{IoU}(\hat{\mathbf{b}}_i,\mathbf{b}).
\end{array}
\end{equation}
Here $\mathcal{B}_c$ is ground-truth boxes of class $c$, $\tau_{reg}=0.5$, and $\omega^{reg}_{ic}$ includes category weights and hard-negative weights.
\begin{algorithm}[t]
\caption{LADTP training for DFAP-UGC}
\label{alg:ladtp}
\begin{algorithmic}[1]
\REQUIRE Training set $\mathcal{D}$, base weights $\lambda^0_t$, smoothing factor $\rho$, exponent $r$
\ENSURE Trained DFAP-UGC model
\STATE Initialize model parameters and task KPI $K_t$.
\FOR{each epoch $e$}
    \STATE Clear the epoch loss buffer.
    \FOR{each mini-batch $(I^r,I^p,\mathcal{Y})$}
        \STATE Predict boxes, categories, severity scores, regional logits, and quality logits.
        \STATE Assign dense positives by task-aligned assignment.
        \STATE Compute $\mathcal{L}_{cls}$, $\mathcal{L}_{bbox}$, $\mathcal{L}_{giou}$, $\mathcal{L}_{score}$, $\mathcal{L}_{reg}$, $\mathcal{L}_{qual}$.
        \STATE Optimize $\sum_t \lambda_t^e\mathcal{L}_t$ and record each raw loss.
    \ENDFOR
    \STATE Convert epoch-average losses to KPI scores and update EMA KPI.
    \STATE Compute DTP weights and the localization factor.
    \STATE Update task weights for the next epoch.
\ENDFOR
\end{algorithmic}
\end{algorithm}
\subsection{Dense-IoU Quality Branch and Inference Ranking}
The dense-IoU quality branch estimates localization reliability, which is different from severity prediction. It takes detached dense query features and detached predicted boxes:
\begin{equation}
\hat{\mathbf{q}}_i=Q([\mathrm{sg}(\mathbf{h}_i);\mathrm{sg}(\hat{\mathbf{b}}_i)]).
\end{equation}
For each class, the training target is the maximum IoU between the predicted box and ground-truth boxes of that class when the IoU is above 0.3; otherwise, the target is zero. This design lets the quality head learn to calibrate candidate ranking without directly reshaping the main localization head.
The dense-IoU quality branch is optimized by
\begin{equation}
\begin{array}{l}
\mathcal{L}_{qual}=\frac{1}{N_{qual}^{+}}\sum_{i,c}\omega^{qual}_{ic}\,\mathrm{BCE}(\sigma(\hat{q}_{ic}),y^{qual}_{ic}),\\
y^{qual}_{ic}=m^{qual}_{ic}\mathbf{1}[m^{qual}_{ic}\ge\tau_{qual}],
m^{qual}_{ic}=\max\limits_{\mathbf{b}\in\mathcal{B}_c}\mathrm{IoU}(\hat{\mathbf{b}}_i,\mathbf{b}).
\end{array}
\end{equation}
where $\tau_{qual}=0.3$ and $\omega^{qual}_{ic}$ down-weights dense negative query-class pairs.

At inference time, DFAP-UGC combines the main classification probability, the regional probability, and the quality probability to compute the confidence score:
\begin{equation}
C_{ic}=\sigma(\hat{p}_{ic})\cdot\sigma(\hat{r}_{ic})\cdot\sigma(\hat{q}_{ic})^{0.5}.
\end{equation}
Each query is decoded in single-label mode by choosing the class with the largest confidence. Class-wise NMS with IoU threshold 0.5 is then applied, and the final output is $\{\hat{\mathbf{b}},c,\hat{s}\}$.
\subsection{LADTP Training}
Although DFAP-UGC relies on dense assignment and ranking branches, UEAP remains a coupled multi-task problem: localization affects whether classification, severity regression, regional verification, and quality estimation receive meaningful supervision. Inspired by Dynamic Task Prioritization (DTP)~\cite{guo2018dynamic}, we propose Locality-Aware DTP (LADTP) training strategy to balance task losses according to epoch-level task difficulty and localization quality.

Let $\mathcal{T}=\{cls,bbox,giou,score,reg,qual\}$ denote the losses. At epoch $e$, LADTP records the average raw loss $\bar{L}^{e}_{t}$ for each task and converts it into a KPI score. For localization-related losses, we use
\begin{equation}
\kappa^{e}_{t}=\exp(-\mathrm{clip}(\bar{L}^{e}_{t},0,2)),\quad t\in\{bbox,giou\},
\end{equation}
and for the remaining losses,
\begin{equation}
\kappa^{e}_{t}=\exp(-\mathrm{clip}(\bar{L}^{e}_{t},0,3)),\quad t\in\{cls,score,reg,qual\}.
\end{equation}
The KPI is smoothed by exponential moving average:
\begin{equation}
K^{e}_{t}=\rho K^{e-1}_{t}+(1-\rho)\kappa^{e}_{t}.
\end{equation}
The dynamic task weight is
\begin{equation}
w^{e}_{t}=-(1-K^{e}_{t})^{r}\log K^{e}_{t}.
\end{equation}
To make the weighting locality-aware, the GIoU KPI produces a localization factor
\begin{equation}
\eta^{e}=\frac{1}{1+\exp[-4(K^{e}_{giou}-1)]}.
\end{equation}
The effective loss weight for the next epoch is
\begin{equation}
\lambda^{e}_{t}=\left\{\begin{array}{ll}
\lambda^{0}_{t}w^{e}_{t}, & t\in\{bbox,giou\},\\
\lambda^{0}_{t}w^{e}_{t}\eta^{e}, & t\in\{cls,score,reg,qual\}.
\end{array}\right.
\end{equation}
The proposed LADTP training method is summarized in Algorithm~\ref{alg:ladtp}.
Thus, when localization is unreliable, LADTP prevents classification, severity, regional, and quality objectives from dominating optimization too early; as localization improves, these tasks receive stronger effective weights.
\begin{table}[t]
  \centering
  \renewcommand{\arraystretch}{1.3}
  \setlength{\tabcolsep}{4.5pt}
  \begin{tabular}{lcccc}
  \hline
  Method & AP$_{50}$ & mAP & R$_{50}$ & MAE \\
  \hline
  DFAP-UGC & \textbf{0.4328} & \textbf{0.2038} & \textbf{0.8106} & \underline{0.3222} \\
  Faster R-CNN-sum & \underline{0.3758} & 0.1719 & 0.6951 & \textbf{0.3215} \\
  Faster R-CNN-single & 0.3755 & 0.1682 & \underline{0.7021} & 0.3303 \\
  Faster R-CNN-concat & 0.3734 & \underline{0.1720} & 0.6999 & 0.3226 \\
  YOLOv12n-concat & 0.3058 & 0.1661 & 0.5848 & 2.0990 \\
  YOLOv12n-single & 0.2813 & 0.1524 & 0.5617 & 2.3276 \\
  YOLOv12n-add & 0.2783 & 0.1485 & 0.5702 & 2.6422 \\
  YOLOv1-concat & 0.2044 & 0.0771 & 0.4492 & 0.3407 \\
  YOLOv1-single & 0.1932 & 0.0738 & 0.4438 & 0.3385 \\
  YOLOv1-sum & 0.1897 & 0.0710 & 0.4394 & 0.3354 \\
  \hline
  \end{tabular}
  \caption{Test comparison between DFAP-UGC and the adapted baselines, including Faster R-CNN, YOLOv1/v12n.}
  \label{tab:main_results}
\end{table}
\section{Experiments}
We first present experimental settings, followed by performance results, ablation studies, and a case study.
\subsection{Experimental Setup}
We evaluate all methods on UEAP-4k, which contains 4,222 paired samples. 
The dataset is split into 3,040 training images, 338 validation images, and an untouched test set of 844 images with 2,729 ground-truth anomaly boxes. 
A prediction is counted as a true positive only when its category is correct and its IoU with an unmatched ground-truth box is at least 0.5. 
We report AP at IoU threshold 0.5 (AP$_{50}$), mAP over IoU thresholds 0.50:0.05:0.95, Recall$_{50}$ (R$_{50}$), and severity score MAE on matched true positives. 
All reported DFAP-UGC detection metrics use the original ranked detections without confidence-threshold filtering.
We train DFAP-UGC for 50 epochs using AdamW with learning rate $5\times10^{-5}$, backbone learning rate $5\times10^{-6}$, weight decay $10^{-4}$, and the total batch size is set as 8 during training and inference. 
\subsection{Comparison with Adapted Baselines}
Since existing detectors cannot directly solve UEAP with paired images and severity scores, we adapt Faster R-CNN~\cite{ren2015faster}, YOLOv12n~\cite{tian2026yolov12}, and YOLOv1~\cite{redmon2016you} to the same output protocol.
For pairwise inputs, both reference and enhanced images are passed through the backbone, and their features are fused by concatenation or summation before predicting anomaly boxes, categories, and severity scores.
We also report the single-image setting, which uses only the enhanced image.
Table~\ref{tab:main_results} compares DFAP-UGC with the adapted baselines on the test set.
DFAP-UGC achieves the best AP$_{50}$, mAP, and R$_{50}$, reaching 0.4328 AP$_{50}$, 0.2038 mAP, and 0.8106 R$_{50}$.
Compared with the strongest Faster R-CNN baseline, DFAP-UGC improves AP$_{50}$ by 5.70 points and mAP by 3.18 points, while also substantially improving recall.
Faster R-CNN variants are competitive in severity MAE, but their AP and recall are consistently lower.
YOLOv12n and YOLOv1 variants underperform in AP and recall, suggesting that one-stage grid prediction struggles with small and fine-grained enhancement artifacts.
Overall, the comparison shows that explicitly modeling problem-reference differences with dense DETR-style queries provides stronger detection quality than adapting standard detectors.
\begin{table}[t]
  \centering
  \renewcommand{\arraystretch}{1.3}
  \setlength{\tabcolsep}{4.5pt}
  \begin{tabular}{lcccc}
  \hline
  Variant & AP$_{50}$ & mAP & R$_{50}$ & MAE \\
  \hline
  Full DFAP-UGC & \underline{0.4328} & \textbf{0.2038} & \textbf{0.8106} & \underline{0.3222} \\
  w/o dynamic weighting & \textbf{0.4370} & \underline{0.1991} & 0.7999 & 0.3291 \\
  w/o quality head & \underline{0.4328} & 0.1937 & 0.8010 & 0.3317 \\
  w/o TAL & 0.3914 & 0.1620 & \underline{0.8080} & 0.3326 \\
  Problem-only fusion & 0.4235 & 0.1934 & 0.8069 & \textbf{0.3142} \\
  w/o regional classifier & 0.4119 & 0.1839 & 0.8054 & 0.3332 \\
  w/o regional HN & 0.4260 & 0.1961 & 0.7893 & 0.3306 \\
  w/o class-weighted box & 0.4253 & 0.1923 & 0.7948 & 0.3357 \\
  w/o box-aware quality & 0.4233 & 0.1911 & 0.7963 & 0.3244 \\
  \hline
  \end{tabular}
  \caption{
  Ablation results of DFAP-UGC variants on the test set.
  Each variant removes or weakens one component from the full model.
  ``w/o TAL'' replaces task-aligned assignment with dense focal supervision;
  ``Problem-only fusion'' removes reference and difference features;
  ``w/o regional HN'' disables regional hard-negative mining;
  and ``w/o box-aware quality'' removes predicted-box input from the quality head.
  }
  \label{tab:ablation}
\end{table}
\begin{figure*}[th]
\centering
\includegraphics[width=\textwidth]{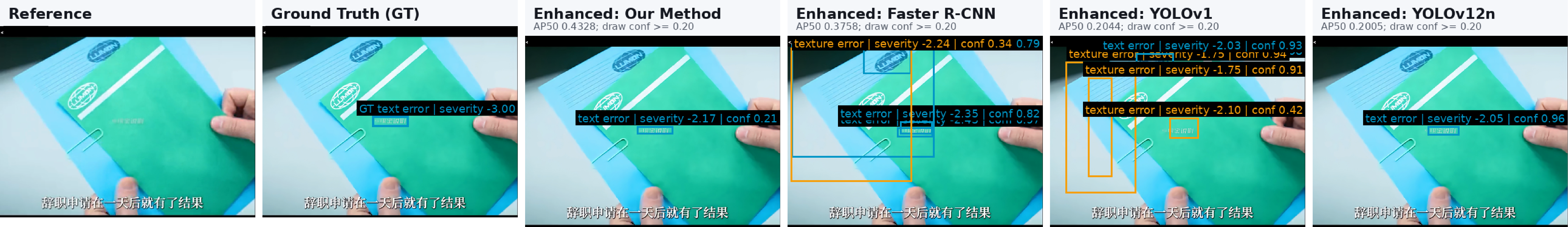}\\[3pt]
\includegraphics[width=\textwidth]{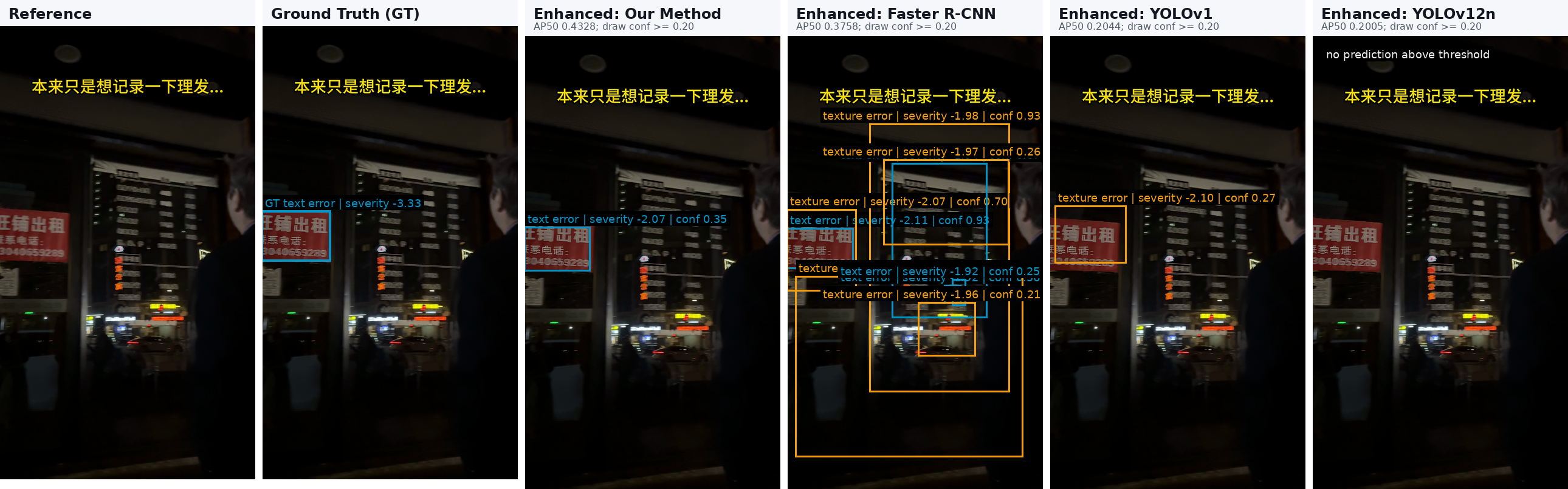}
\caption{Qualitative inference cases on UEAP-4k. From left to right in each case: reference image, ground-truth annotation on the enhanced image, DFAP-UGC prediction, Faster R-CNN prediction, YOLOv1 prediction, and YOLOv12n prediction.}
\label{fig:qualitative_cases}
\end{figure*}
\subsection{Ablation Study}
Table~\ref{tab:ablation} reports ablation results of DFAP-UGC on the test set.
All variants use the original ranked detections without confidence-threshold filtering.
Starting from the full model, each variant removes or weakens one component, including dynamic loss weighting, the dense-IoU quality head, TAL assignment, paired reference/difference fusion, regional classifier, regional hard-negative mining, class-weighted box regression, and box-aware quality estimation.

Full DFAP-UGC tops AP$_{50}$, mAP, and R$_{50}$, ranks second in MAE, and shows the best overall balance. Dynamic weighting improves overall metrics despite a slight AP$_{50}$ drop; removing it reduces mAP, R$_{50}$, and MAE. TAL removal causes the largest AP loss, confirming its key role in dense supervision, while single-image input lowers AP, validating paired reference and difference features. All ablations contribute positively, yet none match the full model's mAP and recall.

\subsection{Inference Case Study}
Figure~\ref{fig:qualitative_cases} presents some representative inference cases from UEAP-4k. Each row compares the reference image, the ground-truth annotation on the enhanced image, DFAP-UGC prediction, YOLO prediction, and Faster R-CNN prediction. Compared with the adapted detectors, DFAP-UGC produces predictions that are more consistent with the annotated local anomalies, especially for subtle enhancement-induced distortions that require comparing the enhanced image with a non-ideal reference. These cases show that the proposed pairwise difference-fusion design improves local anomaly perception while reducing missed or noisy detections in UGC scenes. 
\section{Conclusion}
We introduced UEAP, a new paired-image anomaly perception task for identifying local failures introduced by UGC enhancement. Based on this task, we constructed UEAP-4k and proposed DFAP-UGC, which combines explicit problem-reference difference fusion with dense spatial queries, regional verification, quality-aware ranking, and locality-aware dynamic task prioritization. Experiments show that DFAP-UGC achieves stronger performances than the adapted detectors reported in our comparison. The ablation study validates the contribution of the main training and ranking designs.

\bibliography{aaai2027}


\end{document}